\documentclass[sigconf]{acmart}

\AtBeginDocument{%
  }

\usepackage{multirow}

\begin{document}

\title[Logic-Free Building Automation with Deep Learning (Preprint version)]{Logic-Free Building Automation: Learning the Control\\of Room Facilities with Wall Switches and Ceiling Camera}

\author{Hideya Ochiai}
\email{ochiai@g.ecc.u-tokyo.ac.jp}
\orcid{0000-0002-4568-6726}
\affiliation{%
  \institution{The University of Tokyo}
  \country{Japan}
}

\author{Kohki Hashimoto}
\email{k-84mo10@g.ecc.u-tokyo.ac.jp}
\affiliation{%
  \institution{The University of Tokyo}
  \country{Japan}
}

\author{Takuya Sakamoto}
\email{takuya50719@g.ecc.u-tokyo.ac.jp}
\affiliation{%
  \institution{The University of Tokyo}
  \country{Japan}
}

\author{Seiya Watanabe}
\email{watanabe-seiya441@g.ecc.u-tokyo.ac.jp}
\affiliation{%
  \institution{The University of Tokyo}
  \country{Japan}
}

\author{Ryosuke Hara}
\email{hara-ryosuke573@g.ecc.u-tokyo.ac.jp}
\affiliation{%
   \institution{The University of Tokyo}
  \country{Japan}
}

\author{Ryo Yagi}
\email{yagi-ryo143@g.ecc.u-tokyo.ac.jp}
\affiliation{%
   \institution{The University of Tokyo}
  \country{Japan}
}

\author{Yuji Aizono}
\email{aizono-yuji238@g.ecc.u-tokyo.ac.jp}
\affiliation{%
  \institution{The University of Tokyo}
  \country{Japan}
}

\author{Hiroshi Esaki}
\email{hiroshi@wide.ad.jp}
\affiliation{%
  \institution{The University of Tokyo}
  \country{Japan}
}



\begin{abstract}
Artificial intelligence enables smarter control in building automation by its learning capability of users' preferences on facility control. Reinforcement learning (RL) was one of the approaches to this, but it has many challenges in real-world implementations. We propose a new architecture for logic-free building automation (LFBA) that leverages deep learning (DL) to control room facilities without predefined logic. Our approach differs from RL in that it uses wall switches as supervised signals and a ceiling camera to monitor the environment, allowing the DL model to learn users' preferred controls directly from the scenes and switch states. This LFBA system is tested by our testbed with various conditions and user activities. The results demonstrate the efficacy, achieving 93\%-98\% control accuracy with VGG, outperforming other DL models such as Vision Transformer and ResNet. This indicates that LFBA can achieve smarter and more user-friendly control by learning from the observable scenes and user interactions.
\end{abstract}





\setcopyright{none}
\acmConference[ ]{ }{ }{ }
\acmBooktitle{ }
\acmPrice{ }
\acmDOI{ }
\acmISBN{ }

\renewcommand\footnotetextcopyrightpermission[1]{}

\maketitle

\section{Introduction}

\begin{figure*}
\centering
\includegraphics[width=0.92\textwidth]{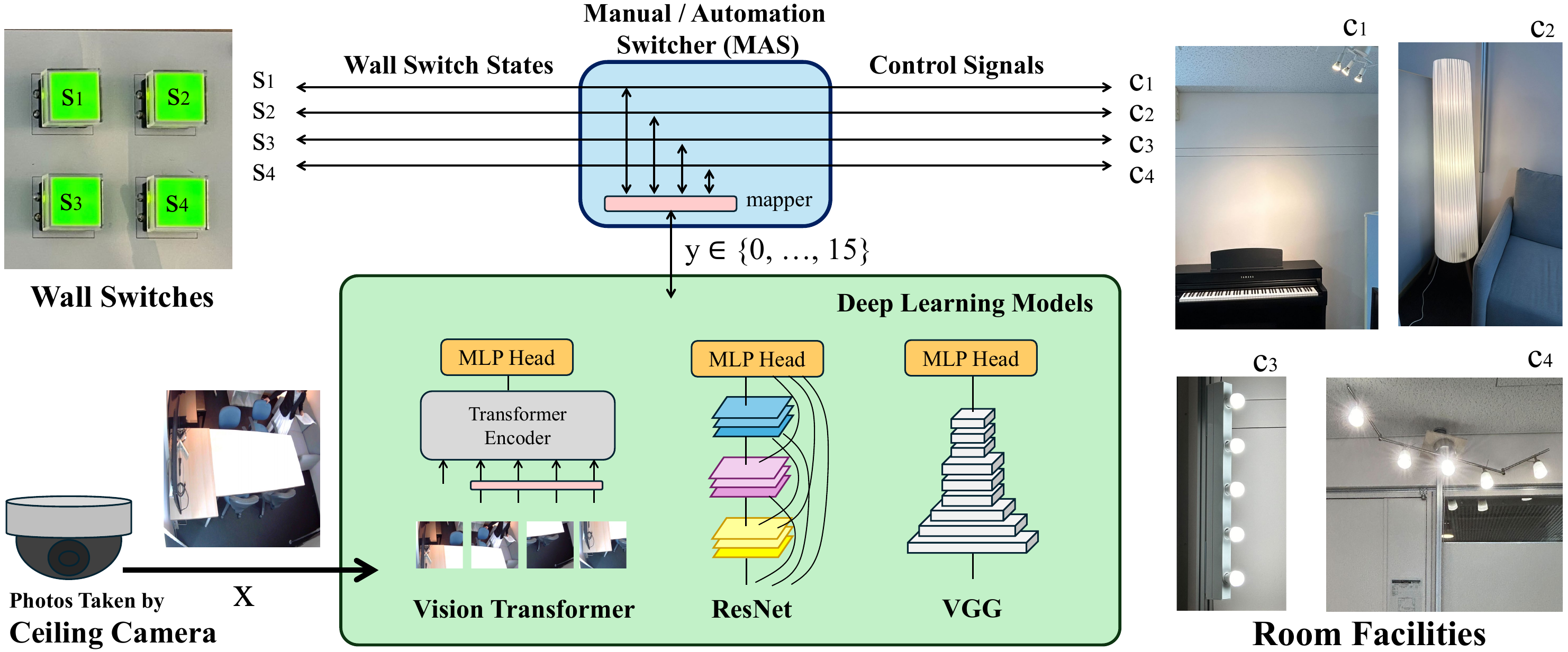}
\caption{Logic-Free Building Automation with Deep Learning. This architecture trains a DL model using the wall switch states and the scene obtained from the ceiling camera. In automation mode, the model directly controls the associated facilities.}
\label{fig:Architecture}
\end{figure*}

Recent advancements in artificial intelligence (AI) enable smarter control in building automation (BA). The traditional BA also showed its smartness but was ``programmed smart''. Now, the learning capability of AI may perform \textbf{program-less BA}, enabling an innovative smarter control for the spaces.

Program-less control of room facilities is challenging because machines do not know the user's preferred control. Reinforcement learning (RL) \cite{yu2021review, wang2020reinforcement, biemann2021experimental, yu2019deep, xu2020multi, jimenez2021sinergym} was one of the approaches for program-less control but encountered many difficulties in practice when applying to the real world\cite{dulac2020empirical}.

We propose logic-free building automation (LFBA) -- a new architecture of facility control with deep learning (DL). LFBA uses (1) a ceiling camera as the sensor for monitoring the space, and (2) wall switches for both manual control and supervising the control. We introduce a manual-automation switcher (MAS) for managing the data flow among wall switches, a DL model, and facilities. 

DL models such as Vision Transformer (ViT) \cite{dosovitskiy2020image}, ResNet\cite{he2016deep}, and VGG\cite{simonyan2014very} have deep potential in learning the scene of a space: e.g., standing, sitting on a chair or sofa, using monitor, reading books, and playing the piano. We use such capabilities for outputting the facility control signals. This differs from object detection (OD) \cite{redmon2016you, liu2016ssd, carion2020end} based approaches because such systems require the users to implement additional logic for the preferred control.

We developed a real-world testbed in May 2024 for studying LFBA systems, collecting the dataset, and testing the performance in practical scenarios. Although we focused on room light control with a ceiling camera, LFBA could be extended to other facilities in smart homes or buildings such as energy management, HVAC, and other appliances. The use of a microphone and a large language model can be considered as another input modality. However, we left these extensions as future items.

We carried out evaluations with three major DL models (ViT, ResNet, VGG) with 5-fold cross-validation. Surprisingly, we found that VGG, the legacy DL model, outperformed others.

Future buildings will have built-in ceiling cameras for facility control although there are discussions about privacy. Related to this, this study was conducted with an ethics review at our institution.

This is the summary of the contributions:

\begin{itemize}
\item We propose the architecture of LFBA -- as an alternative program-less BA to RL-based BA.

\item We developed a real-world testbed and collected a dataset considering diversity and 5-fold cross-validation. \textit{This will be publicly available after the acceptance.}

\item We found that VGG, the legacy DL model, outperformed other major DL models.

\end{itemize}

This paper is organized as follows. Section 2 addresses the related work. We propose the LFBA in section 3 and present the testbed and dataset in section 4. Section 5 shows our evaluation. Section 6 concludes this work.

\section{Related Work}

Many researchers have tried to apply AI to BA systems \cite{himeur2023ai}. The most straightforward approach is to use RL to learn optimal control from the user's feedback of satisfied or not for control trials. Many researchers have worked on RL in smart buildings \cite{yu2021review, wang2020reinforcement}, HVAC controls \cite{biemann2021experimental}, and smart homes \cite{yu2019deep, xu2020multi}. Although applying RL to BA sounds promising, we encounter many issues in practice:

\begin{enumerate}
\item \textbf{RL needs thousands of user feedback by reward or penalty at every trial of control.} This user interface is bothersome for users. Especially, if RL performed control errors for aggressive training, users will be frustrated.

\item Simulator-assisted training is studied to solve these issues \cite{jimenez2021sinergym}. However, \textbf{the reward function, i.e., user feedback, has to be properly and explicitly formulated to run simulations.} This formulation was actually what we did as control logic programming in the legacy BA systems.
\end{enumerate}

For these reasons, we explore another architecture that does not rely on RL but uses legacy wall switches for learning the user's preferred control.

Other related studies are the use of OD-enabled cameras such as YOLO\cite{redmon2016you}, SSD\cite{liu2016ssd}, and DETR\cite{carion2020end}, for smart homes \cite{liciotti2020sequential, huu2021proposing, mehmood2019object, huang2020development}. However, the users must implement their control logic explicitly for their room. This is an extension of programmed-smart. Programming for every room is challenging for most of the users.

Our focus is the design of the architecture for real-world implementation. The LFBA, we propose in this paper, has the learning capability of the user's preferred control, allowing logic-free and user-friendly building automation.




\section{Logic-Free Building Automation with Deep Learning}

This section defines the architecture of logic-free building automation with deep learning (LFBA). For simplicity, this paper focuses on binary (i.e., 0/1) cases for switch states and control signals. 

\begin{figure*}
\centering
\includegraphics[width=0.90\textwidth]{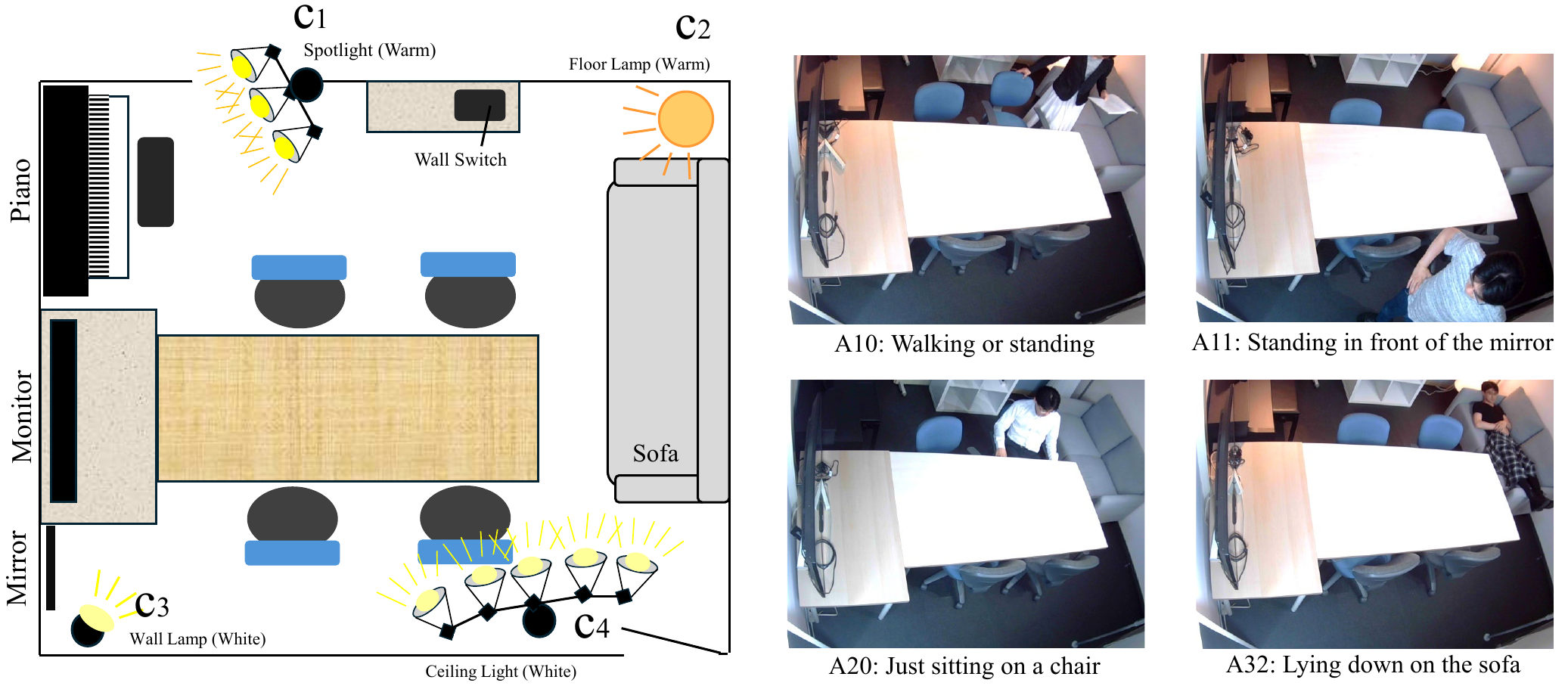}
\caption{The configuration of Logic-Free Building Automation Testbed at our institute. We have collected the dataset by shooting photos from a ceiling camera as Table \ref{tab:dataset_profile} with several lighting conditions and clothing.}
\label{fig:Dataset}
\end{figure*}

\subsection{System Architecture}

Fig. \ref{fig:Architecture} shows the architecture of LFBA. It is organized with wall switches, building facilities (control targets), a ceiling camera, a deep learning model, and a manual-automation switcher (MAS).

Let $s_i$ be the state of a switch, and $c_i$ be the state of the corresponding control signal. Here, $i=1, \ldots, n$ is the index of the switch and the associated facility. In manual mode, $s_i$ alternates when the user physically pushes the switch. For example, $s_i$ becomes 1 after pushing switch $i$ at $s_i=0$. If $c_i=1$, the corresponding facility is active. We denote $\textbf{s}=(s_1, \ldots, s_n)$ and $\textbf{c}=(c_1, \ldots, c_n)$ in short.

Regarding the DL model, we denote the output by label $y$ and the input by image $\textbf{x}$. Here, $y$ represents $2^n$ classes -- the combinations of $\textbf{s}$. Image $\textbf{x}$ is taken periodically, e.g., every second, and pushed into the model. The final layer, i.e., multi-layer perceptron (MLP) head, is replaced to match $2^n$ classes. We consider the case where the room does not have more than 10 switches. The case of a larger number should be studied in the future.

\subsection{Manual-Automation Switcher}

This switcher mechanism is important in training the DL model and automating the facility control. MAS has a static 1-to-1 mapper $\textbf{c}=map(y)$, which generates control signals $\textbf{c}$ from model output $y$. To generate $y$ from switch states $\textbf{s}$ for model training, we denote by $y=map^{-1}(\textbf{s})$.

MAS manages the modes of training, prediction, and none. We define the following three operation modes.

\begin{description}
\item[1. Manual (No Training):] The user sets MAS to this mode when he/she does not want to train or use the model. In this mode, the switcher bypasses as $\textbf{c} \leftarrow \textbf{s}$. 

\item[2. Manual (with Training):] The user sets MAS to this mode when he/she wants to train the model. In this mode, the switcher bypasses as $\textbf{c} \leftarrow \textbf{s}$, and uses $\hat{y} \leftarrow map^{-1}(\textbf{s})$ as a supervised label. The DL model trains its parameters with $\textbf{x}$ shot by the ceiling camera and $\hat{y}$. 

\item[3. Automation:] The user sets MAP to this mode when he/she wants to rely on the DL model for automation. In this mode, the model generates $y$ from image $\textbf{x}$ periodically. The output is used to control building facilities. $\textbf{c} \leftarrow map(y)$. 

\end{description}

\section{LFBA Testbed and Dataset}

We set up a research testbed at our institute in May 2024 and captured the scenes of the room on this testbed.

Fig. \ref{fig:Dataset} shows the structure of the target room and photo examples shot by a ceiling camera. People work, study, have discussions, relax, check their looks, and play the piano. Four controllable lights ($c_1, \ldots, c_4$) are deployed, whose status can be controlled manually by the wall switches or automatically by the output of a DL model.

Table \ref{tab:dataset_profile} shows the dataset profile. We have assigned an \textbf{ID} for scene class. The \textbf{Shots} indicates the number of images for the scene class. The \textbf{Output} denotes the user's preferred control (i.e., ground-truth label), ordering from $c_1$ to $c_4$. For example, at A41 (About to start playing the piano), ``1000'' means $c_1$ should be active, whereas the others should be inactive. In this case, only the spotlight $c_1$ is activated (see Figs. 1 and 2). At this moment, we manually assigned the \textbf{Output} labels to the collected images. 

\begin{figure}
\centering

  \begin{minipage}[b]{0.18\linewidth}
  \centering
  \includegraphics[width=1.0\linewidth, height=115px]{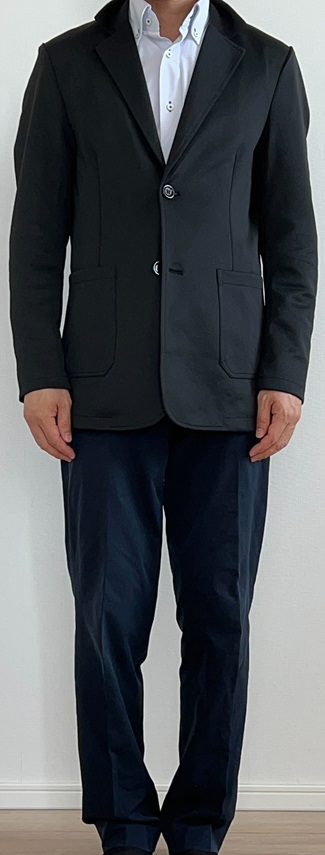}\\
  Run 1\\ 3517 shots
  \end{minipage}
  \begin{minipage}[b]{0.18\linewidth}
  \centering
  \includegraphics[width=1.0\linewidth, height=115px]{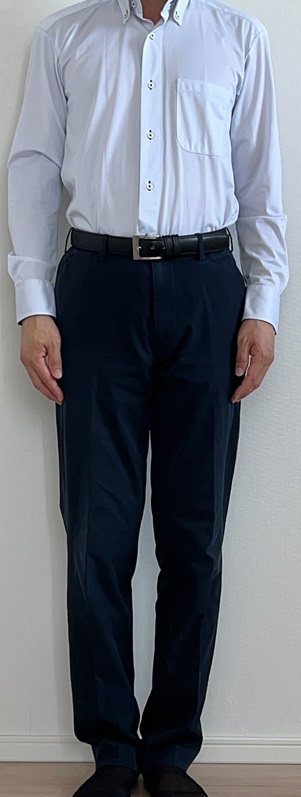}\\
  Run 2\\ 3206 shots
  \end{minipage}
  \begin{minipage}[b]{0.18\linewidth}
  \centering
  \includegraphics[width=1.0\linewidth, height=115px]{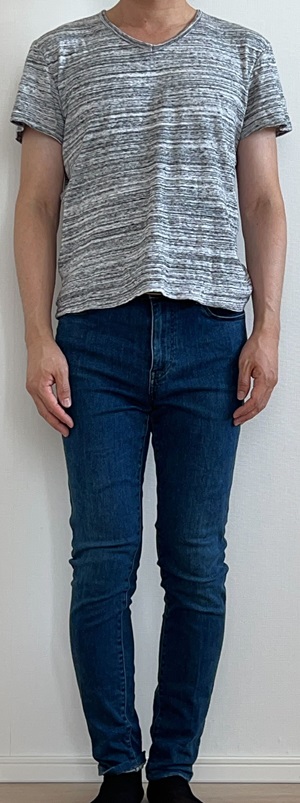}\\
  Run 3\\ 3437 shots
  \end{minipage}
  \begin{minipage}[b]{0.18\linewidth}
  \centering
  \includegraphics[width=1.0\linewidth, height=115px]{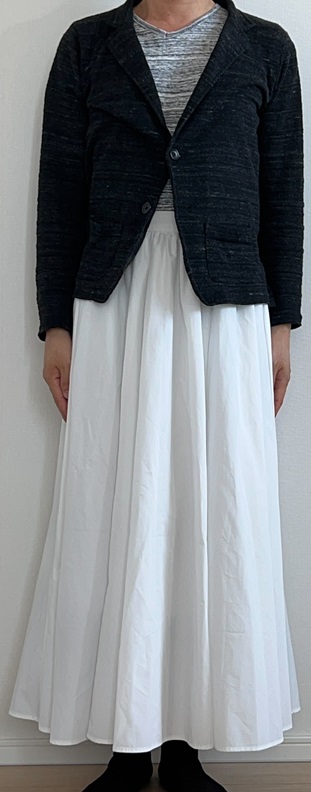}\\
  Run 4\\ 3773 shots
  \end{minipage}
  \begin{minipage}[b]{0.18\linewidth}
  \centering
  \includegraphics[width=1.0\linewidth, height=115px]{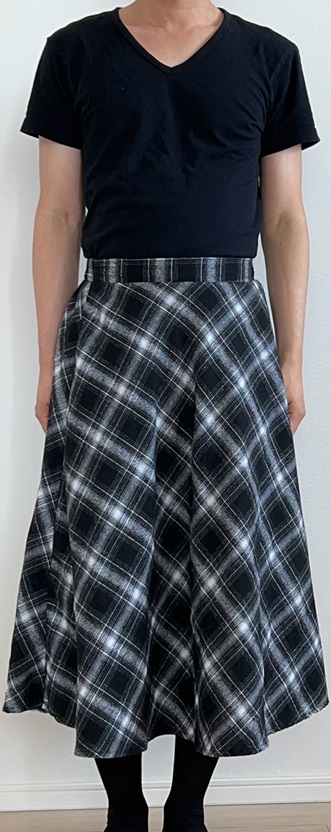}\\
  Run 5\\ 3905 shots
  \end{minipage}

\caption{Data collection runs by different clothes for diversity. Separation by runs allows effective 5-fold cross-run validation.}
\label{fig:DatasetRun}
\end{figure}

To increase the \textbf{diversity} and to make 5-fold cross-validation effective, we performed with different clothing as Fig. \ref{fig:DatasetRun}. The shooting round was separated by what we call ``run''. Hairstyles and facial features could be also extracted at the tuning phase of the DL model, but at this moment, we only focused on the clothing because it would be the major part of the user's appearance. 

\textit{The dataset and related codes for evaluation will be publicly available soon.}



\begin{table}[!t]
\caption{The profile of the dataset. Output indicates the user's preferred control: ordered as $c_1c_2c_3c_4$. Shots indicates the number of images. We collected 17838 images totally. }
\begin{center}
    \begin{tabular}{c|l|c|r}
        \hline \hline
        \textbf{ID} & \textbf{Scene} & \textbf{Output} & \textbf{Shots} \\
        \hline
          A00  &  No one is there. & 0000 & 1468\\
        \hline
          A10  &  Walking or standing. & 0010 & 3567\\
          A11  &  Standing in front of the mirror. & 1111 & 344\\
        \hline
          A20  &  Just sitting on a chair. & 0010 & 1309\\
          A21  &  Sitting on a chair reading papers. & 0011 & 1890 \\
          A22  &  Sitting on a chair with a computer. & 0011 & 2306 \\
          A23  &  Sitting on a chair using a monitor. & 0001 & 2170\\
        \hline
          A30  &  Just sitting on the sofa. & 0110 & 1091 \\
          A31  &  Sitting on the sofa reading papers. & 0100 & 1366 \\
          A32  &  Lying down on the sofa. & 0000 & 971 \\
        \hline
          A40  &  Sitting in front of the piano. & 1010 & 474 \\
          A41  &  About to start playing the piano. & 1000 & 333 \\
          A42  &  Playing the piano. & 1000 & 549 \\
        \hline \hline
    \end{tabular}
\label{tab:dataset_profile}
\end{center}
\end{table}


\section{Evaluation}

We conducted evaluations mainly focusing on comparing the accuracies among major DL models. Further analyses are left open for future work. We set two evaluation cases as follows.
\begin{description}
\item[Merge \& Split Runs:] Evaluation by merging all the runs, shuffling, and splitting into train and test by the ratio of 80\% and 20\%. In this case, clothing for the test also appears in the training data.

\item[5-Fold Cross Run:] Evaluation by picking up a shooting run for a test, and using the other runs for training. In this case, the clothing in the test does not appear in the training data. We performed 5-fold cross-validation by changing the pick-up run.
\end{description}

We calculate balanced accuracy (B-Acc) and the standard accuracy (S-Acc). B-Acc is more suitable as the number of shots by label is unbalanced (see Table \ref{tab:dataset_profile}).

\subsection{Experiment Setting}

We used ViTs, ResNets, and VGGs pre-trained with ImageNet-1K. Please note that ViTs are known as state-of-the-art at many image recognition tasks, ResNets are CNN-based high-performance models, and VGGs are legacy deep learning models.

We replaced the final fully connected layer to fit the output to 16 classes. We tuned the whole model parameters with the following hyperparameters. 

\begin{itemize}
\item Cross entropy loss
\item SGD with momentum = 0.9
\item Batch size = 10
\item Learning rate = 0.001
\item Training epoch = 25
\end{itemize}

We did not apply data augmentation such as random rotation and resizing because the positions in the image are important.

\subsection{Performance of Control Generation}

Table \ref{tab:balanced_accuracy} shows the results. \textbf{In Merge and Split}, all the models achieved almost the similar performances at 98\% (S-Acc) and 97-98\% (B-Acc). \textbf{In 5-Fold Cross Run}, surprisingly, VGG -- the legacy deep learning model performed the best, achieving 94.3\% 
 (S-Acc) and 93.1\% (B-Acc). These accuracy scores show the potential efficacy of LFBA.

Table \ref{tab:balanced_accuracy} shows that ViT could not perform the highest accuracy for unknown clothing, but VGG could. The reason might be that ViTs paid more attention to individual features such as clothing and gave wrong outputs. Of course, further investigations, for example using explainable AI, must be required but we left it open for the future.

\section{Conclusion}

We proposed the architecture of logic-free building automation with deep learning and developed a real-world testbed. Our evaluation results, i.e., 93\%-98\% accuracy, demonstrate LFBA's efficacy as a new building automation architecture.

\begin{table}[!t]
\caption{The performances of control signal generation in merge\&split runs and 5-fold cross run. S-Acc stands for standard accuracy and B-Acc stands for balanced accuracy.}
\label{tab:balanced_accuracy}
\begin{center}
    \begin{tabular}{c|cc|cc}
        \hline \hline
        \multirow{2}{*}{\textbf{Model}}& \multicolumn{2}{c|}{\textbf{Merge \& Split Runs}} & \multicolumn{2}{c}{\textbf{5-Fold Cross Run}}\\
        \cline{2-5}
         & \textbf{S-Acc} & \textbf{B-Acc} & \textbf{S-Acc} & \textbf{B-Acc} \\
        \hline
        
        ViT-B/16 & 0.978 & 0.969 & 0.861 & 0.792 \\
        ViT-L/16 & 0.976 & 0.967 & 0.890 & 0.837 \\
        ResNet-18 & 0.980 & 0.975 & 0.915 & 0.875 \\
        ResNet-50 & 0.980 & 0.974 & 0.934 & 0.915 \\
        ResNet-152 & \textbf{0.983} & \textbf{0.977} & 0.939 & 0.920 \\
        VGG-16 (BN) & \textbf{0.983} & \textbf{0.977} & 0.938 & 0.921 \\
        VGG-19 (BN) & 0.981 & \textbf{0.977} & \textbf{0.943} & \textbf{0.931} \\
        \hline \hline

    \end{tabular}
\end{center}
\end{table}

%
%
%
%
%
%
%
%
%
%
%
%
%
%
%
%
%


\bibliographystyle{ACM-Reference-Format} 
\bibliography{main}

\end{document}